\documentclass{article} % For LaTeX2e
\usepackage{iclr2025_conference,times}

\usepackage{amsmath,amsfonts,bm}

\def\eqref#1{equation~\ref{#1}}
\def\1{\bm{1}}

\DeclareMathAlphabet{\mathsfit}{\encodingdefault}{\sfdefault}{m}{sl}
\SetMathAlphabet{\mathsfit}{bold}{\encodingdefault}{\sfdefault}{bx}{n}

\usepackage{times}
\usepackage{soul}
\usepackage{url}
\usepackage[hidelinks]{hyperref}
\usepackage[utf8]{inputenc}
\usepackage[small]{caption}
\usepackage{graphicx}
\usepackage{amsmath}
\usepackage{amsthm}
\usepackage{booktabs}
\usepackage{algorithm}
\usepackage{algorithmic}
\usepackage[switch]{lineno}
\usepackage{algorithm}
\usepackage{algorithmic}
\usepackage{xcolor}
\usepackage{subcaption}
\usepackage{diagbox}
\usepackage{cancel}
\usepackage{hyperref}
\usepackage{makecell}

\title{Human-like compositional learning of visually-grounded concepts using synthetic environments}

\author{
\textbf{Zijun Lin}\textsuperscript{1,3} \quad
\textbf{M Ganesh Kumar}\textsuperscript{2,3} \quad
\textbf{Cheston Tan}\textsuperscript{3} \\
\textsuperscript{1}Nanyang Technological University \\
\textsuperscript{2}Harvard University \\
\textsuperscript{3}Centre for Frontier AI Research, A*STAR
}

\iclrfinalcopy
\begin{document}

\maketitle

\begin{abstract}
    The compositional structure of language enables humans to decompose complex phrases and map them to novel visual concepts, showcasing flexible intelligence. While several algorithms exhibit compositionality, they fail to elucidate how humans learn to compose concept classes and ground visual cues through trial-and-error. To investigate this multi-modal learning challenge, we designed a 3D synthetic environment in which an agent learns, via reinforcement, to navigate to a target specified by a natural language instruction. These instructions comprise nouns, attributes, and critically, determiners, prepositions, or both. The vast array of word combinations heightens the compositional complexity of the visual grounding task, as navigating to a blue cube above red spheres is not rewarded when the instruction specifies navigating to ``some blue cubes below the red sphere''. We first demonstrate that reinforcement learning agents can ground determiner concepts to visual targets but struggle with more complex prepositional concepts. Second, we show that curriculum learning—a strategy humans employ—enhances concept learning efficiency, reducing the required training episodes by 15\% in determiner environments and enabling agents to easily learn prepositional concepts. Finally, we establish that agents trained on determiner or prepositional concepts can decompose held-out test instructions and rapidly adapt their navigation policies to unseen visual object combinations. Leveraging synthetic environments, our findings demonstrate that multi-modal reinforcement learning agents can achieve compositional understanding of complex concept classes and highlight the efficacy of human-like learning strategies in improving artificial systems' learning efficiency.
\end{abstract}

\section{Introduction}
Humans describe the world by combining linguistic parts-of-speech (concept classes) such as adjectives and nouns. As visual complexity increases, we use more words~\citep{sun2022seeing} and additional concept classes to enhance description accuracy \citep{elsner2018visual,qiao2020referring}. For example, determiners are essential for describing object quantity and ownership \citep{lee2023determinet}, while prepositions are critical for expressing spatial and temporal relationships between objects \citep{agrawal2023stupd}. In contrast, pretrained vision-language models struggle to ground visual objects to complex linguistic concepts and flexibly compose this knowledge to solve novel scenarios \citep{kamath2023s, shen2023scaling}. This gap restricts the alignment between human and machine communication for collaboration.

% Children master the use of determiner and preposition concept classes by the age of three \citep{abu2004describing,brown1973first}. They gradually learn to apply these classes to describe novel situations \citep{tomasello1987learning,washington1978children}, generalizing beyond their learned experiences. With sufficient exposure, children begin to compose across concept classes to convey greater detail in their descriptions \citep{valian1986syntactic,gleason2022development}. For instance, they can accurately describe novel visual scenes such as ``some red balls above the green box'' by combining determiners, prepositions, nouns, and adjectives. Altering any word or their sequence can significantly change the meaning. Such generalization abilities are lacking in current state-of-the-art vision-language models \citep{okawa2024compositional}, even when trained on datasets far larger than a child's experience.

Children typically master the use of determiners and prepositions by age three~\citep{abu2004describing, brown1973first}, applying these concepts to describe novel situations and generalizing beyond their experiences~\citep{tomasello1987learning, washington1978children}. They learn incrementally, beginning with simple terms and advancing to complex concepts~\citep{valian1986syntactic, gleason2022development, richards1984language}. The ability to compose linguistic elements for detailed descriptions, such as ``some red balls above the green box,'' is a skill even advanced vision-language models lack~\citep{okawa2024compositional}. Tailoring learning curricula to individual pace and knowledge remains an ongoing challenge, both for humans and machine learning systems~\citep{bengio2009curriculum, graves2008language, soviany2022curriculum, wang2021survey, narvekar2020curriculum}.

% Moreover, children learn individual concepts incrementally, following a loose curriculum \citep{bengio2009curriculum} that progresses from simple concepts like above'' and below'' to more complex ones such as ``alongside'' \citep{gleason2022development,richards1984language}. Crucially, the curriculum must be tailored to a child's learning pace and prior knowledge, as a slower-paced curriculum does not necessarily improve learning outcomes \citep{graves2008language}. Determining the optimal curriculum for both humans and machine learning models remains an open challenge \citep{soviany2022curriculum,wang2021survey,narvekar2020curriculum}.

In this work, we developed several 3D synthetic environments to study the visuo-linguistic grounding problem in a reinforcement learning setting, similar to how children could learn these concepts. A trial-and-error based learning environment facilitates artificial agents to learn from experience, a requisite to continually improve alignment during human-machine interaction. Specifically, we investigate the influence of a curriculum in improving compositional learning to ground determiner and preposition concept classes to visual objects. \textbf{Notably, our synthetic environments allow precise control over the generation of training and held-out test sets while ensuring balanced data across classes, a challenge with real-world datasets.} Leveraging on synthetic environments, we make the following contributions:

\begin{itemize}
\item We demonstrate that reinforcement learning agents, when trained naively to maximize rewards, can ground descriptive attributes (determiners) to single objects but struggle to ground relational attributes between objects (prepositions).
\item We show that agents cannot generalize determiners and prepositions to new combinations without incremental curriculum training from simple to complex tasks.
\item We establish that agents can reason compositionally about novel I.I.D. and O.O.D. test instructions and navigate to unknown visual objects after learning determiner and preposition concepts.
\end{itemize}

%Furthermore, they mostly learn through by trial-and-error and with sparse teaching signals instead of supervised learning methods.

%Unfortunately, many architectures do not facilitate learning compositional structures. They learn representations to directly solve the training task but not a representational structure that can directly solve an unseen combination. This problem can either be solved using architectures that enforce learning compositional structures or generating datasets that vary the composition of information so that models learn the structure.

%Here, we focus on the latter. 

% \begin{itemize}
%     \item Humans describe the real world by summing up parts of language. This makes it easy to decompose instructions and quickly understand the semantics. This is the benefit of compositional structures. 
%     \item However, deep learning models do not learn compositional representations or structures. Although some approaches like contrastive learning allows models to learn to cluster information, their ability to generalize the compositional understanding to new combinations is not evaluated, especially in a RL setting
%     \item Furthermore, humans learn to compose information in a myriad of learning strategies e.g. supervised, self-supervised and reinforcement learning. How the latter does it is unclear. 
%     \item 
% \end{itemize}

\section{Synthetic Environments for Grounding}

This section describes the synthetic environments designed for agents to learn to ground visual objects to determiner and preposition linguistic concepts by trial-and-error learning.

\subsection{Environment Design} 

\begin{figure}
\centering
\includegraphics[width=1\textwidth]{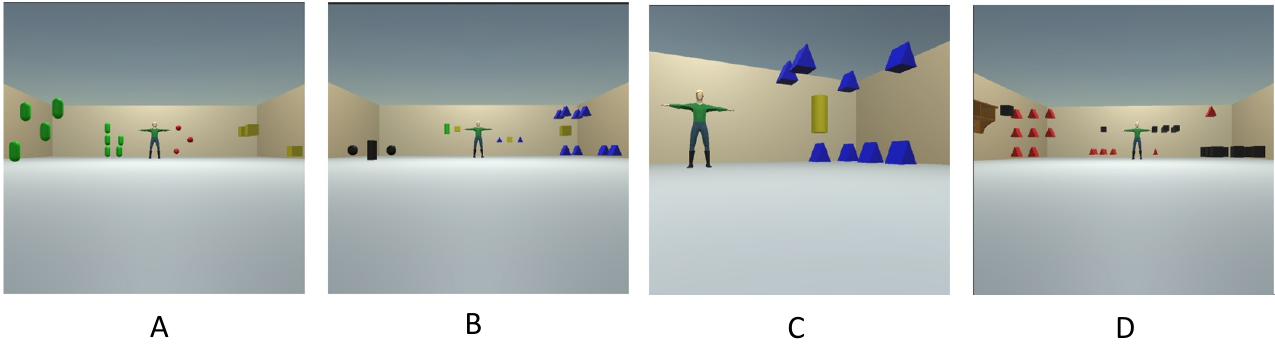}
\caption{Example environments, each with four target options. A reward of +10 is given when the agent navigates to the target matching the instruction. Punishments of -1, -3, and -10 are incurred for hitting a wall, reaching the wrong target, or failing to reach the correct target, respectively. \textbf{A}: Agent's view of the determiner ($D$) environment. \textbf{B}: Agent's view of the preposition ($P$) environment. \textbf{C}: Close-up view of the target object in $P$. \textbf{D}: Agent's view of the combined determiner and preposition ($D+P$) environment.}
\label{fig:env}
\end{figure}

As shown in Fig. \ref{fig:env}, we built three synthetic 3D environments to enable the agents to learn four concepts --- shape, color, determiner and preposition. While our primary emphasis is on determiner and preposition, the agents also need to acquire knowledge of color and shape throughout the learning process. Specifically, objects come in random colors (red, green, blue, yellow and black) and shapes (capsule, cube, cylinder, prism and sphere), each chosen from a set of 5.

The target object is randomly placed in one of four predetermined locations within a rectangular room, as illustrated in Fig. \ref{fig:env}. Additionally, the positions of the individual objects within each location are randomized within a 3×3×3 block. This creates numerous visual representations that are constrained by the description of the instruction. The room layout remains constant and includes fixed visual landmarks like a door, window, shelf, and a reference man. The first-person perspective is simulated through a Unity-based camera, capturing the environmental dynamics in RGB images. These images serve as the visual input for the agent. Additionally, the environment generates textual instructions describing the target object, which are provided to the agent as language input. Details of each environment follows below.

\subsection{Determiner Environment}

In the determiner environment ($D$), agents learn eight determiners: ``A'', ``Few'', ``Some'', ``Many'', ``This'', ``That'', ``These'', and ``Those''. Instructions follow the form ``Determiner $+$ Object(s)''. For example, ``\emph{few} yellow cubes'' or ``\emph{those} green cylinders'' (Fig. \ref{fig:env}A). The first four determiners describe object quantity: ``A'' denotes a single object, 2--3 objects as ``Few'', 4--6 as ``Some'', and 7--9 as ``Many''. The last four determiners depend on proximity to a reference man (Fig. \ref{fig:env}A): ``This'' and ``These'' refer to closer objects (single or multiple 2$\sim$9), while ``That'' and ``Those'' refer to farther objects.

Non-targets are designed to test grounding: (1) same determiner, random color/shape; (2) same color/shape, random determiner; (3) random determiner, color, and shape. For example, given the target instruction ``\emph{Many} red cylinder'', non-target visual objects could be ``\textit{Few} red cylinder'', ``Many \textit{blue cube}'', and ``\textit{A green prism}''.

\subsection{Preposition Environment}

In the preposition environment ($P$), agents learn eight prepositions: ``Above'', ``Below'', ``In front of'', ``Behind'', ``Beside'', ``On'', ``Between'', and ``Among''. Instructions follow the form ``Object A $+$ Preposition $+$ Object B'', where Object A and Object B differ in color and shape. For the first six prepositions, Object A and Object B are single instances (e.g., ``green capsule \emph{On} red cube''). For ``Between'' and ``Among'', Object A is singular, while Object B consists of 2--9 instances, respectively.

Non-targets are designed similarly: (1) same preposition, random color/shape; (2) same color/shape, random preposition; (3) random preposition, color, and shape. For example, given the target instruction ``green cube \emph{Above} blue cylinder'', non-target visual objects could be ``green cube \textit{In front of} blue cylinder'', ``\textit{yellow cube} Above \textit{black capsule}'', and ``\textit{red cylinder Among green cube}''.

\subsection{Combined Determiner and Preposition Environment}

The combined environment ($D+P$) integrates determiners and prepositions. Instructions follow the form ``Determiner A $+$ Object A $+$ Preposition $+$ Determiner B $+$ Object B''. For example, ``a red cube Above Many black cubes'' (Fig. \ref{fig:env}D).

This environment evaluates the influence of curriculum learning and the ability to decompose complex instructions in few-shot. Non-targets are designed to test comprehension of all concepts: (1) swapped color/shape attributes; (2) altered determiners; (3) modified prepositions. For example, given the target instruction ``A black cube Above Many red prisms'', non-target visual objects could be A ``\textit{red prism} Above Many \textit{black cubes}'', ``\textit{Many} black cubes Above \textit{A} red prisms'', and ``A black cube \textit{Behind} Many red prisms''.

\subsection{Evaluating compositional learning and generalization}

In each episode, the agent receives specific rewards based on its actions. Navigating to the target object yields a reward of +10, while collisions with non-target objects or walls result in penalties of -3 and -1, respectively. Additionally, the agent incurs a penalty of -10 if it fails to reach the target within the maximum allowed steps ($T_{max}=500$). We define the \textit{Performance Criterion} as the agent achieving a perfect reward of +10 in at least 800 out of 1000 episodes, corresponding to an 80\% success rate. The number of training episodes required to meet this criterion in the determiner and preposition environments is detailed in Section \ref{exp1}.

Table \ref{tab:determiner table} and Table \ref{tab:preposition table} (supplementary material) show the train-test split setting for the $D$ and $P$ environments respectively. After training the agents on the training combinations, the agent's were evaluated on its zero-shot compositional generalization ability on the held-out test combinations. The results are in  Section \ref{exp2}. Specifically, $D$ comprises 200 unique instructions (160 train, 40 test), while $P$ comprises a total of 6000 unique instructions (4800 train, 1200 test). 

Naively training agents to learn the $D+P$ environment is intractable as $D+P$ contains 160,000 unique combinations (120,000 train, 40,000 test), which is 800 and 27 times more combinations than the $D$ and $P$ environments respectively. Hence, we used the $D+P$ environment to evaluate agents' ability to decompose novel determiner-preposition instructions. This is by first training the agents on the determiner $D$ and prepositions $P$ environment separately, until they reached performance criterion, and subsequently train them in the $D+P$ environment for only 100,000 episodes. 

\section{Agent Architecture}

\begin{figure}[t]
  \centering
\includegraphics[width=1\linewidth]{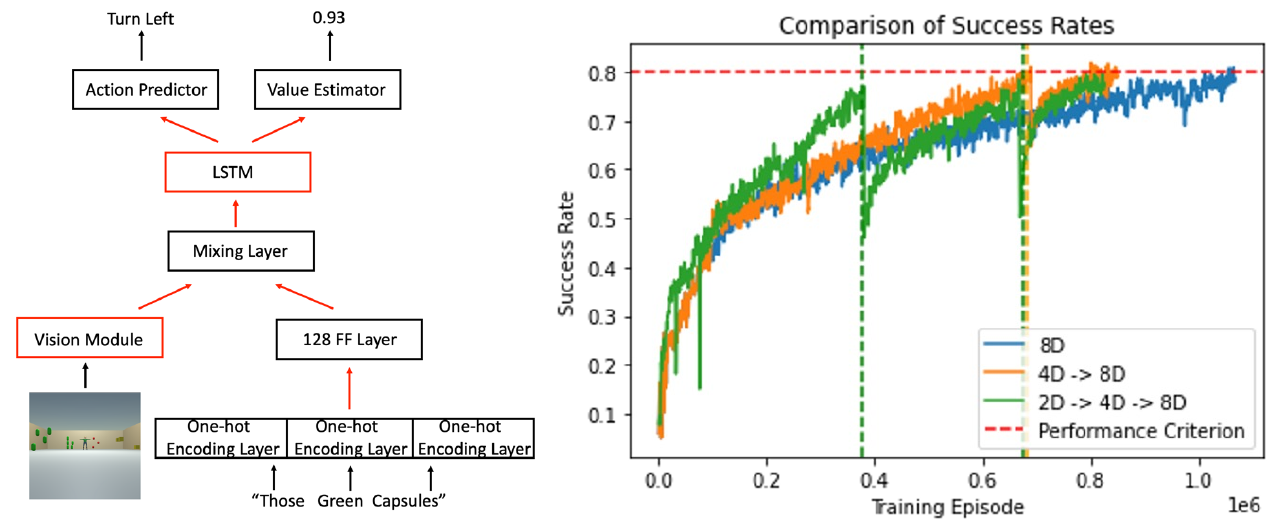}
\caption{\textbf{Left}: Agent architecture. \textbf{\textcolor{red}{Red}} arrows or boxes represent \textbf{\textcolor{red}{trainable}} weights, while \textbf{\textcolor{black}{black}} arrows or boxes represent \textbf{\textcolor{black}{frozen}} weights. \textbf{Right}: Success rates of the agents with and without curriculum learning in $D$.}
\label{fig:agent}
\end{figure}

% Fig. \ref{fig:agent} (left) shows the vision-language agent architecture adapted from \citep{Hill, lin2023compositional}, which takes in visual inputs (a 3$\times$128$\times$128 tensor of RGB pixel values) and language input (one-hot vector embedding) to output a probability distribution $\pi(a_t|s_t)$ over four possible actions, i.e., move forward, move backward, turn left and turn right. Details of the agent architecture are described in Section \ref{sec:arch} (supplementary material). 

% The agent's objective is to maximize cumulative discounted reward \citep{sutton2018reinforcement, kumar2024model} by navigating to the correct visual object based on specific language instruction \citep{wang2019reinforced, kumar2022nonlinear} while avoiding incorrect objects or walls, which incur negative rewards. Training is performed using the advantage actor-critic (A2C) algorithm \citep{mnih2016asynchronous,kumar2021onea}, optimized with the RMSProp optimizer at a fixed learning rate of $2.5\times10^{-4}$ across all experiments.

Our vision-language agent architecture, adapted from \citet{Hill, lin2023compositional}, processes visual inputs (a 3$\times$128$\times$128 tensor of RGB pixel values) through a vision module consisting of three convolutional layers. In the Determiner environment ($D$), language instructions are encoded as three one-hot vectors representing the Determiner, Color, and Shape components. For the Preposition environment ($P$), five one-hot vectors are used, while in the combined environment ($D+P$), seven one-hot vectors represent the instructions. These are passed to a linear layer, then concatenated with the vision module's output to form an embedding, which is processed by a Long Short Term Memory (LSTM) module, whose activity $s_t$ informs both the action predictor (actor) and value estimator (critic). The actor derives a probability distribution $\pi(a_t|s_t)$ over the actions (move forward, move backward, turn left, turn right), while the critic estimates the state-value function $V(s_t)$. 

The agent's objective is to maximize cumulative discounted rewards~\citep{sutton2018reinforcement, kumar2024model} by navigating to correct visual targets based on language instructions~\citep{wang2019reinforced, kumar2022nonlinear}, while avoiding incorrect options incurring penalties. Training leverages the advantage actor-critic (A2C) algorithm~\citep{mnih2016asynchronous,kumar2021onea}, optimized with RMSProp at a learning rate of $2.5\times10^{-4}$ across all experiments.

\section{Experiments and Results}

This section evaluates the effects of a curriculum on compositional learning and generalization in grounding determiner and preposition concepts to visual objects. In experiment 1, we investigate the role of using curriculum to improve learning efficiency. Here, the agent has to ground either eight determiners or prepositions to visual objects in the Determiners or Preposition environments. 

Experiment 2 examines the I.I.D. and O.O.D. generalization capabilities of agents. Agents' I.I.D. generalization capabilities are examined on held-out test instruction combinations in either the Determiners ($D$) or Preposition ($P$) environments in zero-shot (no training). O.O.D. generalization is evaluated by using agents pretrained on the Preposition ($P$) environments, to solve the Determiner and Preposition environment ($D+P$) in few-shot training. This demonstrates whether agents can rapidly adapt to an environment that contains a significantly greater number of potential combinations of instructions.

% Agents' I.I.D. zero-shot capabilities are examined on held-out test instruction combinations in either the Determiners ($D$) or Preposition ($P$) environments. While agents have seen all the determiner or preposition instructions, they were trained on a specific determiner/preposition - shape combination Table \ref{tab:determiner table}, \ref{tab:preposition table}. 

% pretrained in the Preposition ($P$) environments in the Determiner and Preposition environment ($D+P$). This demonstrates whether agents can rapidly adapt to an environment that contains a significantly greater number of potential combinations of instructions. Specifically, the agent needs to break down the held-out test instruction and navigate to novel visual objects. 

\subsection{Experiment 1: Curriculum improves learning efficiency}
\label{exp1}

\begin{table}[t]
    \centering
    \scriptsize
    \begin{tabular}{|c|c|c|}
        \hline
        \multicolumn{1}{|c|}{\textbf{Determiner}} & \multicolumn{1}{c|}{\centering\textbf{Training}} & \multicolumn{1}{c|}{\centering\textbf{Total}} \\
            \multicolumn{1}{|c|}{\textbf{Environment}} & \multicolumn{1}{c|}{\centering\textbf{Episodes (M)}} & \multicolumn{1}{c|}{\centering\textbf{Episodes (M)}} \\
        \hline
        \textbf{$8D$} & $0.87$  & $0.87$ \\
        \hline
         \textbf{$4D\rightarrow8D$} & $0.66\rightarrow0.11$ & $0.77$ \\
        \hline
        \textbf{$2D\rightarrow4D\rightarrow8D$} & $0.38\rightarrow0.33\rightarrow0.11$ & $0.82$\\
        \hline
    \end{tabular}
    \caption{Curriculum learning ($4P \rightarrow 8P$ and $2P \rightarrow 4P \rightarrow 8P$) reduces the total number of episodes needed to learn the Determiner ($D$) environment, compared to without ($8P$) a curriculum. Values in the table indicate the number of episodes (in \textbf{millions}) needed to achieve a success rate $\geq$ 80\% (performance criterion) over 1000 episodes. Lower values indicate faster learning. }
    \label{tab: determiner result}
\end{table}

\begin{table}[t]
    \centering
    \scriptsize
    \begin{tabular}{|c|c|c|}
        \hline
        \multicolumn{1}{|c|}{\textbf{Preposition}} & \multicolumn{1}{c|}{\centering\textbf{Training}} & \multicolumn{1}{c|}{\centering\textbf{Total}} \\
            \multicolumn{1}{|c|}{\textbf{Environment}} & \multicolumn{1}{c|}{\centering\textbf{Episodes (M)}} & \multicolumn{1}{c|}{\centering\textbf{Episodes (M)}} \\
        \hline
        \textbf{$8P$} & $>3.5$  & $>3.5$ \\
        \hline
         \textbf{$4P\rightarrow8P$} & $1.57\rightarrow1.3$ & $2.87$ \\
        \hline
        \textbf{$2P\rightarrow4P\rightarrow8P$} & $0.9\rightarrow0.3\rightarrow0.98$ & $2.18$\\
        \hline
    \end{tabular}
    \caption{Curriculum learning ($4P \rightarrow 8P$ and $2P \rightarrow 4P \rightarrow 8P$) reduces the total number of episodes needed to learn the Prepositions ($P$) environment, compared to without ($8P$) a curriculum. Values in the table indicate the number of episodes (in \textbf{millions}) needed to achieve a success rate $\geq$ 80\% (performance criterion) over 1000 episodes. Lower values indicate faster learning. }
    \label{tab:preposition results}
\end{table}

To emulate human-like learning, we employ a curriculum, where agents progressively learn simpler concepts before advancing to more complex ones. For both the determiner and preposition environment, agents are trained under three setups: (1) learning all eight concepts directly ($8D$/$8P$), (2) learning four simple concepts first, then all eight e.g. ($4D\rightarrow8D$), and (3) learning two, then four, and finally all eight e.g. ($2D\rightarrow4D\rightarrow8D$).

The results in Table \ref{tab: determiner result} and Fig. \ref{fig:agent} (right) show that curriculum learning reduces the total training episodes by 15\% in the $4D\rightarrow8D$ setup compared to learning all eight determiners from scratch. While the $2D\rightarrow4D\rightarrow8D$ setup does not save more episodes than $4D\rightarrow8D$, it still achieves a modest reduction of 0.05M episodes. These improvements are significant given the relatively small number of training instruction combinations in the determiner environment (160 combinations). Simple concepts were determined based on the fastest learning curves for each concept. 

In contrast, the preposition environment ($P$) presents a much greater challenge, with 4800 instruction combinations—30 times more than the determiner environment. As shown in Table \ref{tab:preposition results}, agents trained directly on all eight prepositions ($8P$) fail to converge even after 3.5M episodes. This aligns with expectations, as the complexity of the $8P$ environment would require approximately 21M episodes to learn, based on the 0.87M episodes needed for the determiner environment. However, curriculum learning significantly improves learning efficiency: the $4P\rightarrow8P$ and $2P\rightarrow4P\rightarrow8P$ setups reduce the required training episodes to 2.87M and 2.18M, respectively. This demonstrates that curriculum learning is essential for tackling environments with high combinatorial complexity.

\subsection{Experiment 2: Compositional learning for I.I.D. and O.O.D. Generalization}
\label{exp2}

\begin{table}[t]
    \centering
    \scriptsize
    \begin{tabular}{|c|c|}
        \hline
       \textbf{Environment} & \textbf{Success Rate (\%)}  \\
        \hline
        \textbf{$8D$} & $77.5 \pm 0.01$  \\
        \hline
         \textbf{$4D\rightarrow8D$} & $79.0 \pm 0.06$  \\
        \hline
        \textbf{$2D\rightarrow4D\rightarrow8D$} & $76.0\pm 0.24$ \\
        \hline
                 \textbf{$8P$} & $24.8\pm 0.85$  \\
        \hline
         \textbf{$4P\rightarrow8P$} & $76.1\pm 0.99$  \\
        \hline
        \textbf{$2P\rightarrow4P\rightarrow8P$} & $76.4\pm 0.46$ \\
        \hline
    \end{tabular}
    \caption{Zero-shot generalization in $D$ and $P$ on held-out instructions. Values in the table indicate the mean and standard deviation of the success rate of the agents in the test environment over 3 iterations ($N=3$) in percent. The RL agents are tested in the held-out instructions right after reaching the success rate of 80\%, except for $8P$, which does not converge even after 3.5 million training episodes. The closer the results approach 80\%, the better compositionality the agents show.}
    \label{tab: determiner test result}
\end{table}

% Agents' I.I.D. zero-shot capabilities are examined on held-out test instruction combinations in either the Determiners ($D$) or Preposition ($P$) environments. While agents have seen all the determiner or preposition instructions, they were trained on a specific determiner/preposition - shape combination Table \ref{tab:determiner table}, \ref{tab:preposition table}. 

% pretrained in the Preposition ($P$) environments in the Determiner and Preposition environment ($D+P$). This demonstrates whether agents can rapidly adapt to an environment that contains a significantly greater number of potential combinations of instructions. Specifically, the agent needs to break down the held-out test instruction and navigate to novel visual objects. 

We evaluated the I.I.D. zero-shot capabilities of agents on held-out test instruction combinations in the Determiner ($D$) and Preposition ($P$) environments. Although agents were exposed to all determiner or preposition instructions, they were trained only on a subset of determiner/preposition-shape combinations (Tables \ref{tab:determiner table} and \ref{tab:preposition table}) to achieve an 80\% success rate. As shown in Table \ref{tab: determiner test result}, agents demonstrated consistent zero-shot performance on held-out test instructions, with success rates of at least 76\%. In contrast, agents that did not converge during training (e.g., $8P$) exhibited chance-level performance (25\%). These findings indicate that agents have acquired the ability to decompose and recompose concepts, enabling generalization to new I.I.D. instructions and visual targets during held-out tests.

The combined Determiner and Preposition environment ($D+P$) presents a greater challenge, with 160,000 O.O.D. instruction combinations—a scale 25 times larger than the preposition environment and 750 times larger than the determiner environment. To address this complexity, we implemented two strategies: (1) pretraining the agent in the Preposition ($P$) environment using a $4P\rightarrow8P$ curriculum, and (2) fine-tuning the agent for compositional learning in the $D+P$ environment for only \textbf{0.1M episodes}. Without these strategies, agent convergence was unattainable. 

Remarkably, with only 2.97M training episodes, the agent achieved an average reward of 7.2 and a \textbf{success rate of 53\%} over 1,000 testing episodes in the $D+P$ environment, which has 160,000 instruction combinations. This contrasts with naive training in the $8P$ environment with just 4,800 instruction combinations, where agents failed to converge even after 3.5M episodes. This demonstrates that once foundational concepts are learned, agents can rapidly adapt to new combinations and solve increasingly complex environments with significantly fewer episodes. Additionally, this highlights the importance of learning to compose concepts using a curriculum to tackle environments with extreme combinatorial complexity.

\section{Conclusion, limitations and future work}

We developed several 3D synthetic environments to illustrate the impact of curriculum learning on instruction based navigation tasks and demonstrated the compositional capabilities of reinforcement learning agents. Notably, we are the first to showcase the feasibility of grounding RL agents in complex instructions involving determiners and prepositions. Our findings reveal that agents can decompose and recompose instructions, akin to human intelligence, allowing them to effectively solve previously unseen I.I.D. and O.O.D. test cases in zero-shot and few-shot respectively. This work marks a significant step towards aligning human and machine interaction (e.g. collaborative robots, autonomous vehicles), as real-life referring expressions often extend beyond simple word forms or adjective-noun combinations to include numerical references and spatial relationships. The unity scripts and code for the environments and agents will be made publicly available upon acceptance. 

The 3D environments crafted for this study utilized simple geometric shapes, such as ``capsule'' and ``prism'', raising concerns about the model's generalization to more realistic objects~\citep{Hill}. Additionally, the absence of obstacles results in straightforward navigation, suggesting that agents trained in these environments may struggle with more complex navigation scenarios that involve obstacles~\citep{anderson2018vision, gu2022vision, kumar2021oneb}. Future research directions include exploring diverse model architectures and integrating pre-trained text and vision encoders \citep{shah2023lm}. Investigating the optimal combination of determiners or prepositions in the curriculum, such as using 2 or 4 determiners to expedite learning, is also worthwhile. A thorough analysis of the model's representations is expected to provide insights into how concepts are embedded~\citep{kumar2024model, lee2023determinet}, enhancing the model's generalization capabilities across diverse scenarios. Moreover, evaluating whether pretrained large language models can effectively ground visual objects to concept classes using these 3D environments offers another avenue of exploration. This could involve using LLMs to make inferences at each time step~\citep{wang2021survey} or generating code to predict actions based on visual frames~\citep{cloos2024generating}.

\bibliographystyle{iclr2025_conference}

\newpage
\appendix
\section{Agent Architecture}
\label{sec:arch}
For vision input, RGB pixel values are passed into the vision module, which contains three convolutional layers, and the output flattened into a 3136 (a 64$\times$7$\times$7 tensor) dimensional embedding. In $D$, the language module takes in three one-hot vector embedding of the instructions, each representing the Determiner, Color and Shape. For example, an instruction in $D$ such as ``Those Green Capsules'' is represented by three one-hot vectors respectively. Followed by the same rule, instructions in $P$ are represented by five one-hot vectors considering that the instructions contain five words and $D+P$ adopts seven one-hot vectors. These one-hot vectors are fully connected to a 128 unit linear embedding layer. The 3136-D vector from the vision module and the 128-D vector from the language module are concatenated and fed into a 256-D linear mixing layer. 

A Long Short Term Memory (LSTM) module takes the 256-dimensional embeddings as input. Its activity $s_t$ is passed to both the action predictor (actor) and value estimator (critic). The action predictor maps the LSTM's activity to a probability distribution $\pi(a_t|s_t)$ over four possible actions, i.e., move forward, move backward, turn left and turn right. Meanwhile, the value estimator computes a scalar approximation of the agent's state-value function $V(s_t)$.

\section{Environmental Design}
 As shown in Table \ref{tab:determiner table} and \ref{tab:preposition table}, 75\% of color-shape combinations were used to train the agents on the relevant tasks and 25\% held-out test combinations were used to evaluate the agents' zero-shot ability to use the rules learned for unseen combinations. For example, during the training phase, the agents are trained on the instructions such as ``Few blue cylinder'' and ``A black cube''. Once the agents reach the performance criterion in training environment, they are tested whether they could combine the concept ``Few'' and ``black cube'' to accurately navigate to ``Few black cube''.
\begin{table*}[ht]
\centering
\scriptsize
\begin{tabular}{|c|c|c|c|c|c|c|c|c|}
\hline
\textbf{Shape\textbackslash{}Determiner} & \textbf{A} & \textbf{Few} & \textbf{Some} &\textbf{Many} & \textbf{This}&\textbf{That} &\textbf{These} &\textbf{Those}  \\
\hline
\textbf{Capsule} & \textbf{\textbf{\textcolor{blue}{Test}}} & Train & Train & \textbf{\textbf{\textcolor{blue}{Test}}}  & Train & Train & Train & Train \\
\hline
\textbf{Cube} & Train & \textbf{\textbf{\textcolor{blue}{Test}}} & Train & Train & \textbf{\textbf{\textcolor{blue}{Test}}}  & Train & Train & Train \\
\hline
\textbf{Cylinder} & Train & Train & \textbf{\textbf{\textcolor{blue}{Test}}} & Train & Train & \textbf{\textbf{\textcolor{blue}{Test}}}  & Train & Train \\
\hline
\textbf{Prism} & Train & Train & Train & \textbf{\textbf{\textcolor{blue}{Test}}}  & Train & Train & \textbf{\textbf{\textcolor{blue}{Test}}} & Train \\
\hline
\textbf{Sphere} & Train & Train & Train & Train & \textbf{\textbf{\textcolor{blue}{Test}}} & Train & Train & \textbf{\textbf{\textcolor{blue}{Test}}}  \\
\hline
\end{tabular}
\caption{Train-Test split for $D$.}
\label{tab:determiner table}
\end{table*}

\begin{table*}[ht]
\scriptsize
\centering
\begin{tabular}{|c|c|c|c|c|c|c|c|c|}
\hline
\textbf{Shape\textbackslash{}Preposition} & \textbf{Above} & \textbf{Below} & \textbf{In front of} &\textbf{Behind} & \textbf{Beside}&\textbf{On} &\textbf{Between} &\textbf{Among}  \\
\hline
\textbf{Capsule} & \textbf{\textbf{\textcolor{blue}{Test}}} & Train & Train & \textbf{\textbf{\textcolor{blue}{Test}}}  & Train & Train & Train & Train \\
\hline
\textbf{Cube} & Train & \textbf{\textbf{\textcolor{blue}{Test}}} & Train & Train & \textbf{\textbf{\textcolor{blue}{Test}}}  & Train & Train & Train \\
\hline
\textbf{Cylinder} & Train & Train & \textbf{\textbf{\textcolor{blue}{Test}}} & Train & Train & \textbf{\textbf{\textcolor{blue}{Test}}}  & Train & Train \\
\hline
\textbf{Prism} & Train & Train & Train & \textbf{\textbf{\textcolor{blue}{Test}}}  & Train & Train & \textbf{\textbf{\textcolor{blue}{Test}}} & Train \\
\hline
\textbf{Sphere} & Train & Train & Train & Train & \textbf{\textbf{\textcolor{blue}{Test}}} & Train & Train & \textbf{\textbf{\textcolor{blue}{Test}}}  \\
\hline
\end{tabular}
\caption{Train-Test split for $P$.}
\label{tab:preposition table}
\end{table*}
\section{Related Work}

\subsection{Determiner and Preposition}

Large vision-language models, excel in zero-shot classification \citep{clip} and text-to-image generation \citep{tao2023galip}, yet they face documented challenges in object counting \citep{paiss2023countclip} and spatial relationship understanding \citep{lewis:arxiv23}. Despite attempts to address these limitations through additional loss functions or fine-tuning \citep{jiang2023clip}, performance improvements remain modest and fall short of perfection.

In the realm of agent navigation, instructions incorporating determiners (e.g., Few, Many) or prepositions (e.g., Above, Below) have received limited attention. Although these words are prominently used by humans since age three and play a fundamental role in realising human-AI interaction, there is a scarcity of studies in this area. Our research seeks to explore the generalization capabilities of agents trained by reinforcement learning to successfully reach target objects when provided with instructions containing determiners or prepositions. Our environment dataset and analysis aims to provide deeper insights into how agents learn spatial relationships and acquire object counting capabilities in a human-like learning setting.

\subsection{Curriculum Learning for Reinforcement learning}

Curriculum learning can be divided the into two categories namely model-level curriculum learning and data-level curriculum learning \citep{soviany2022curriculum}. The former is to dynamically adjust the model's complexity or structure as it learns, potentially making it more adept at handling progressively challenging tasks. The latter focuses on defining a difficulty criterion for the training data or task. The model begins learning from simpler tasks and gradually progress to more challenging ones. This entails organizing the training dataset in a way that facilitates a smooth transition from easy to difficult examples. The underlying principle is to guide the learning process by presenting the model with tasks of increasing complexity over time.

Inspired by how human learns and the non-convex optimization properties of reinforcement learning, curriculum learning has been proposed to improve policy convergence if the task order is well curated \citep{soviany2022curriculum,wang2021survey,narvekar2020curriculum}. However, optimizing the dataset or task sequence is still not well understood and hence a specific curriculum may not always result in successful learning outcome. 

Hill et al. (2020) demonstrated that the data-level curriculum learning approach reduced the number of training episodes needed for a reinforcement learning agent to ground 40 visual objects to its attribute-noun combination. However, a relatively small task as such still required about 600,000 training episodes for convergence. Furthermore, compositional generalization to held-out instruction was not evaluated. Comparatively, the number of instruction combinations involving either determiners and prepositions in our grounding task are 200 and 4800 respectively. We seek to establish if curriculum learning can expedite the concept grounding process and ensure compositional generalization with a task complexity that are orders of magnitude higher. 

\subsection{Models for Compositional Learning}

Compositional learning involves breaking down information into fundamental elements or concepts and then integrating these elements to address new and unfamiliar combinations in a few-shot or zero-shot setting. \citep{lake2015human,xu2021zero,de2006emergence}. 

Compositional learning has primarily been investigated in the realm of object detection, where visual models are trained on pairs of object and attribute information. These models leverage learned invariances to effectively handle unseen test sets \citep{kato2018compositional,purushwalkam2019task,anwaar2021compositional}. Another approach involves using the loss function to encourage networks to break down information into generalized features \citep{stone2017teaching,tolooshams2020convolutional}. Recent advancements include models that can recognize or parse objects in images using bounding boxes \citep{lee2023determinet} or segmentation masks \citep{kirillov2023segment}, enabling them to tackle novel tasks.

Recent multi-modal models learn to align visual inputs to language inputs \citep{clip,ma2023crepe,yuksekgonul2022and} to solve the task of compositional reasoning \citep{lu2023chameleon} such as Visual-Question Answering (VQA) \citep{johnson2017clevr}, Referring Expressions \citep{lee2023determinet}, or augment images using instructions \citep{gupta2023visual}. There are few models that integrate compositional learning across vision, language, and action domains. An example is to train reinforcement learning agents that are grounded to visual inputs and language queries. Although these agents require millions of training episodes in diverse simulated environments, they demonstrate impressive proficiency in solving instruction-based tasks \citep{team2021open}.

Nevertheless, how these models ground vision-language-action representations for compositional learning, what the individual concepts are, and how these concepts are recomposed to solve novel combinations remains elusive. Only recently, has the compositional generalization abilities of reinforcement learning agents trained to ground visual objects to nouns and attributes have been explored \citep{lin2023compositional}. In this work, we aim to demonstrate the compositionality of agents in the 3D navigation task given the complex language instructions containing determiners and prepositions.

% \begin{table}[h!]
% \centering
% \begin{tabular}{|c|c|c|c|}
% \hline
% \textbf{Environment} & \textbf{\makecell{Training \\ Episodes (M)}} & \textbf{\makecell{Unseen Test \\ Reward (Max)}} & \textbf{\makecell{Unseen Test \\ Success Rate (Chance)}} \\
% \hline
% $4P \rightarrow 8P$ & 2.87 & 8.5 (10) & 76\% (0\%) \\
% \hline
% $D+P$ & 0.1 & 7.2 (10) & 53\% (0\%) \\
% \hline
% \end{tabular}
% \caption{Performance Metrics for Different Environments}
% \label{tab:performance}
% \end{table}

% \begin{table}[h!]
% \centering
% \begin{tabular}{ccc}
% \toprule
% \textbf{Environment} & \textbf{Average Reward (Max)} & \textbf{Success Rate (Chance)} \\
% \midrule
% 4P $\rightarrow$ 8P                    & 8.5 (10)                      & 76\% (0\%)                     \\
% D+P                  & 7.2 (10)                      & 53\% (0\%)                     \\
% \bottomrule
% \end{tabular}
% \caption{Performance Metrics for Different Environments}
% \label{tab:performance}
% \end{table}

\end{document}